\documentclass[conference]{IEEEtran}
\IEEEoverridecommandlockouts

\usepackage{url}

\usepackage{breakurl}
\usepackage[breaklinks]{hyperref}

\usepackage[papersize={8.5in,11in},
            top=1in,
            left=0.75in,
            right=0.75in,
            bottom=0.75in]{geometry}

\makeatletter
\def\endthebibliography{%
	\def\@noitemerr{\@latex@warning{Empty `thebibliography' environment}}%
	\endlist
}
\makeatother

\usepackage{stfloats}
\usepackage{tabularx,booktabs}
\usepackage{cite}
\usepackage{authblk}
\usepackage{amsmath,amssymb,amsfonts}
\usepackage{algorithmic}
\usepackage{graphicx}
\usepackage{textcomp}
\usepackage{xcolor}
\usepackage{multirow}
\usepackage{makecell}
\def\BibTeX{{\rm B\kern-.05em{\sc i\kern-.025em b}\kern-.08em
    T\kern-.1667em\lower.7ex\hbox{E}\kern-.125emX}}
\begin{document}

\title{Beyond Simulation: Benchmarking World Models for Planning and Causality in Autonomous Driving  \\
}

\author[1, 2]{Hunter Schofield}
\author[1]{Mohammed Elmahgiubi}
\author[1]{Kasra Rezaee}
\author[2]{Jinjun Shan}

\affil[1]{Noah's Ark Lab, Huawei Technologies Canada}
\affil[2]{York University, Toronto, Canada}
\affil[ ]{\textit {\{hunter.schofield,mohammed.elmahgiubi,kasra.rezaee\}@huawei.com}}
\affil[ ]{\textit {\{hunterls,jjshan\}@yorku.ca}}


\maketitle

\begin{abstract}
\noindent
World models have become increasingly popular in acting as learned traffic simulators. Recent work has explored replacing traditional traffic simulators with world models for policy training. In this work, we explore the robustness of existing metrics to evaluate world models as traffic simulators to see if the same metrics are suitable for evaluating a world model as a pseudo-environment for policy training. Specifically, we analyze the metametric employed by the Waymo Open Sim-Agents Challenge (WOSAC) and compare world model predictions on standard scenarios where the agents are fully or partially controlled by the world model (partial replay). Furthermore, since we are interested in evaluating the ego action-conditioned world model, we extend the standard WOSAC evaluation domain to include agents that are causal to the ego vehicle. Our evaluations reveal a significant number of scenarios where top-ranking models perform well under no perturbation but fail when the ego agent is forced to replay the original trajectory. To address these cases, we propose new metrics to highlight the sensitivity of world models to uncontrollable objects and evaluate the performance of world models as pseudo-environments for policy training and analyze some state-of-the-art world models under these new metrics.     
\end{abstract}

\section{Introduction}
\noindent
Deep learning has become a popular approach for both autonomous vehicle trajectory planning \cite{GameFormer, PDM, NuPlan2nd} and traffic simulation \cite{Trafficbots, SMART}. Traditionally, planning models are either trained using large datasets \cite{WOMD, nuplan}, or in a realistic simulator \cite{CARLA, SMARTS, metadrive}. Datasets provide good baselines to train a model on since they contain observations collected from real-world scenarios, however, due to the nature of data collection, vehicles can only replay their trajectory which results in more aggressive behavior as they cannot deviate from their original route. Simulators on the other hand can be reactive, however, the distribution gap \cite{sim2real} between observations in the simulated environment and the real-world environment makes it difficult for models trained in a simulator to perform well on real hardware. To address this problem, recent works have considered whether using learned simulators trained on real-world data can improve policy training for an individual agent \cite{GUMP}. However, it is still unclear if simulation models that are optimized to perform well on traffic simulation benchmarks such as WOSAC \cite{WOSAC} are also suitable for training an individual driving policy. \\

\begin{figure}[!ht]
    \centering
    \includegraphics[width=\columnwidth]{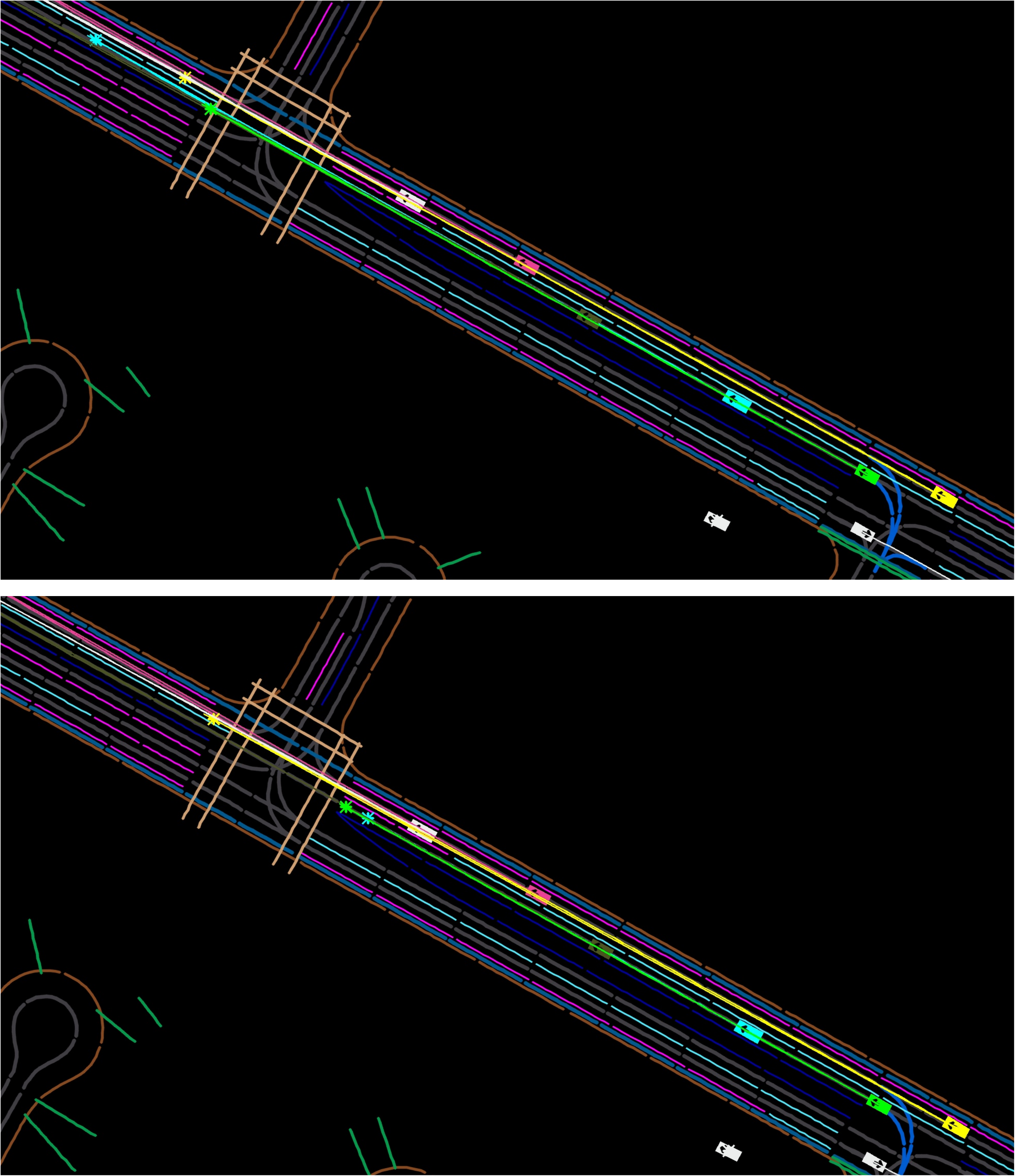}
    \caption{This birds-eye-view image of a WOMD scene represents a motivating example. In this figure, simulations are displayed from TrafficBots V1.5 \cite{Trafficbots} where the model can control all agents (top), and where the model can control all agents except for the ego, which is forced to replay it's original trajectory (bottom). Perturbing the scene by adding an uncontrollable agent creates a conflict between the cyan ego vehicle and the green vehicle behind it as indicated by the close proximity of there final locations indicated by the stars.}
    \label{fig:sim_agent_compare}
\end{figure}

\noindent
In this work, we consider new benchmark metrics for evaluating data-driven learned simulators as closed-loop policy training platforms. We propose a causal agent evaluation domain to explore a model's robustness in predicting realistic futures for agents that are causal to the ego vehicle in a scenario. Furthermore, we propose a reactivity metric that evaluates a model's ability to simulate realistic futures when it is unable to control all objects in the scenario. Extensive evaluations were performed to explore which metrics best indicate a simulation model's ability to enable a good closed-loop training environment. Finally, we explore methods for tuning simulation models to improve their robustness in acting as a closed-loop training environment. \\

\noindent
Our experiments show that existing simulation models are biased towards scenarios where all scenario agents are controllable. Figure 1 depicts a scenario where all objects are simulated (left) and the same scenario where all objects are simulated except for the ego vehicle which is forced to replay its trajectory (right). In the full simulation example, the ego vehicle is predicted to pass through the intersection without slowing down. In the partial simulation, the ego vehicle replay stops at the intersection when the light turns yellow, however, the vehicle behind does not react and collides with the ego. Such a failure on the part of the simulator can significantly degrade the performance of an agent that is trying to minimize a cost associated with collisions as it incentivizes not stopping at intersections. \\

\noindent
The contributions of this work are listed as follows:
\vspace{-1mm}
\begin{itemize}
    \item We propose two new metrics as extensions of the WOSAC metametric to evaluate the performance of a traffic-simulating world model as a closed-loop training environment for policy training.  
    \item We evaluate the robustness of existing traffic-simulating world models and show that they are biased towards scenarios where all vehicles are controllable by the world model. Furthermore, we find that a model that obtains a high metametric, may not be well suited as an environment for training a policy.
    \item We propose control dropout, a fine-tuning method for improving world models' ability to adapt to scenarios where only some objects are controllable, allowing for more realistic traffic simulations, allowing for better training environments for non-world-model agent policies.
\end{itemize} 

\noindent
This is the first work to explore the robustness of traffic-simulating world models when only part of a scenario is controllable. This is a crucial area to investigate as there is growing interest in using data-driven traffic simulations to facilitate policy training \cite{DataDrivenSim}. Benchmarking and improving the robustness of traffic simulating world models will in turn improve policies that are trained inside world model-based data-driven simulators. The paper is organized as follows. Section \ref{s2} introduces related work. Section \ref{s3} outlines the newly proposed metrics for evaluating world models and introduces control dropout for improving world model robustness. Section \ref{s4} explores the experiments conducted to analyze the sensitivity of existing world models to uncontrollable objects and looks at the effects of training with control dropout. Section \ref{s5} provides some discussion to interpret the previous results.  Section \ref{s6} overviews the conclusions of this work.

\section{Related Work \label{s2}}
\subsection{World Models}
\noindent
World models are learned dynamics models that are action-conditioned on the observer. By learning a stochastic transition function that can predict future observations from pairs of prior observations and actions, world models are capable of predicting multiple likely rollouts of the future. These rollouts can be used for traffic simulation and motion prediction \cite{Trafficbots, GUMP, SMART, KIGRAS, VRD}, motion planning \cite{MILE}, and policy learning \cite{Dream2Control, DreamerV2}. In this work, we explore how to evaluate traffic simulation world models as pseudo-environments to facilitate policy learning for a single autonomous driving planning agent. Specifically, we explore if the metrics that are commonly used to evaluate traffic simulation world models, such as the WOSAC metametrics \cite{WOSAC}, are also suitable for evaluating a world model as a pseudo-environment for policy learning.

\subsection{Motion Planning}
\noindent
Many approaches to learning-based motion planning use imitation learning (IL) \cite{GameFormer, Zhao_TNT} to try and replicate expert trajectories, or reinforcement learning (RL) \cite{SimRL, DQNPlanning} to learn a cost function which when minimized produces good driving behavior. More recently, approaches combining both IL and RL \cite{ImitationNotEnough, HREIL} have succeeded as they can quickly learn a good policy through IL, and then further refine the policy using the RL objective. However, both IL and RL have their respective problems. IL suffers from the distribution shift problem \cite{DAGGER} where the model fails when input observations are out of the distribution of the training data. RL suffers from the sim 2 real gap problem \cite{sim2real} where models that are trained in a simulator do not perform well in the real world due to differences between the simulated and real observations. Our work aims to evaluate whether world models can help solve this problem by exploring the implications of having separate policies interact with the world model traffic prediction.


\subsection{Causal Reasoning}
\noindent
Causal reasoning and awareness in machine learning has been growing in popularity in recent years \cite{CausalRepresentation, CausalAnnotations, CausalRepresentationDomainAdaption, CausalOOD}. A causal agent is defined as an agent whose presence directly affects the behavior of the ego agent. Many studies explore the effects of causality in motion forecasting for human trajectory prediction \cite{OODCausal, Sim2RealCausal}. Recent work has shown that many autonomous driving motion forecasting models are overly sensitive to non-causal objects in the environment \cite{CausalAgents}. As we are interested in evaluating world models as data-driven simulators for policy training, it is crucial that we consider the agents that are causal to the ego vehicle in our work. Thus, we take a critical look at the evaluated objects in the WOSAC challenge and propose an extended domain for evaluating the long-term traffic simulations of causal agents.

\section{Methodology \label{s3}}
\noindent
In this section, we will outline the definition of our newly proposed metrics and ways to improve them as well as the domains on which they are evaluated. To clarify new notation, the superscript \textit{sim} is used when considering all agents except for the ego agent. Furthermore, the subscript \textit{eval} will be used to indicate when only the set of WOSAC evaluation agents is considered and the subscript \textit{causal} will be used to indicate when only the set of WOSAC causal agents to the ego vehicle are considered.

\subsection{Policy Aware Metametric}
\noindent
A world model for traffic simulation can be factorized into the ego policy component and the traffic simulator component, following the definition used in WOSAC.

\begin{equation}
    q^{world}(o_t|o^c_{< t}, a_{t-1}) = \pi(a_{t-1}|o^c_{< t}) q(o^{sim}_t|o^c_{< t}) 
\end{equation}

\noindent
As we are ultimately interested in training a policy inside the traffic simulation, we use the action conditioned definition of the world model where $\pi(a_{t-1}|o^c_{< t})$ is the current policy being trained, $q(o^{sim}_t|o^c_{< t})$ is the traffic simulation, $o_t$ is the observation at timestep $t$, and $o^c_{< t} = \begin{bmatrix} o^{map}, o^{signals}, o_{-H-1}, ..., o_{t-1} \end{bmatrix}$ is the combined observation over some history, $H$. \\

\noindent
To evaluate the quality of a world model, the negative log-likelihood (NLL) of real-world samples under the predicted agent distribution is computed for 9 specific metrics: speed, acceleration, angular speed, angular acceleration, distance to the nearest object, collisions, time to collision, distance to the nearest road edge, and road departures.

\begin{equation}
    NLL^m_{q^{world}} = - \frac{1}{|\mathcal{D}|} \sum^{|\mathcal{D}|}\log q^{world}(o^m|o^c_{< t}, a_{t-1})
\end{equation}

\noindent
where $m$ is a specific metric, and $o^m$ is the relevant observation component for computing metric $m$. Each metric is then aggregated into a realism score, $\mathcal{M}$.

\begin{equation}
    \mathcal{M} = \frac{1}{NM} \sum_{i=1}^N \sum_{j=1}^M w_j m_{i, j}
\end{equation}

\noindent
where $M$ is the number of metrics, $N$ is the number of scenarios, and $w$ is the weight of a particular metric. While this aggregate metric is useful for evaluating long-term futures, it fails to capture the contribution made by each agent. Thus, if the aggregate metric is low for a particular scenario rollout, it is indeterminable if the policy or if the world model is at fault. \\

\noindent
To address this limitation, we propose generating a second set of world rollouts, $\hat{q}^{world}$ that are conditioned on the ground truth (GT) policy, $\hat{\pi}$, such that only the ego vehicle has its trajectory replayed. Then, we compute four sets of NLLs that evaluate the original world model, $NLL_{q^{world}}$, the GT conditioned world model, $NLL_{\hat{q}^{world}}$, the factorized simulation of the original world model, $NLL_{q^{sim}}$, and the factorized simulation of the GT conditioned world model $NLL_{\hat{q}^{sim}}$. Using these sets of NLLs, we propose two new metrics, $\Delta \mathcal{M}_i$ and $\Delta \mathcal{M}^{sim}_i$ defined for a specific scenario, $i$.

\begin{equation}
    \Delta \mathcal{M}_i = \mathcal{M}_i - \hat{\mathcal{M}}_i
    \label{eq:delta_metric}
\end{equation}

\begin{equation}
    \Delta \mathcal{M}^{sim}_i = \mathcal{M}^{sim}_i - \hat{\mathcal{M}}^{sim}_i
    \label{eq:delta_metric_sim}
\end{equation}

\noindent
These two delta metrics give more insight into the performance of the factorized policy-traffic simulator world model. In our results, we find that forcing the ego vehicle to replay its trajectory can cause $\Delta \mathcal{M}_i$ to shift in either direction. Thus to appropriately aggregate the delta metrics, we consider the absolute shift.

\begin{equation}
    \Delta \mathcal{M} = \frac{1}{N} \sum_{i=1}^{N} |\mathcal{M}_i - \hat{\mathcal{M}}_i|
\end{equation}

\noindent
The bidirectional shift in $\Delta \mathcal{M}$ provides insight into the performance of both the world model simulator and the ego policy in a scenario. In the ideal case, $\Delta \mathcal{M} \approx 0$ indicates that there is little change in the simulation after replacing the ego vehicle with the replay policy. However, if $\Delta \mathcal{M}_i > \tau$ then the simulation for scenario $i$ is worse when using the GT policy for some arbitrary threshold, $\tau$, suggesting that the traffic simulation is being confused when it is unable to control all agents. It is expected that $\hat{\mathcal{M}}_i > \mathcal{M}_i$ since $\hat{\mathcal{M}}_i$ contains the exact ground truth distributions for the ego vehicle. This is the reason for considering $\Delta \mathcal{M}^{sim}_i$, which evaluates the same rollouts for all agents except for the ego vehicle, thereby removing the ground truth distribution bias from the evaluation induced by $\hat{q}^{world}$. By analyzing both $\Delta \mathcal{M}_i$ and $\Delta \mathcal{M}_i^{sim}$ together, we can see that if $\Delta \mathcal{M}_i > \tau$ and $\Delta \mathcal{M}_i^{sim} < \tau$ then the replay policy causes the traffic simulation to have a worse interaction (such as a collision) with the ego vehicle. Using the new delta metrics, we can define a simulation confusion rate, $C_s$, and policy confusion rate, $C_p$, as follows.

\begin{equation}
    C_s = \frac{|\Delta \mathcal{M} > \tau \cup \Delta \mathcal{M}^{sim} > \tau|}{N}
\end{equation}

\begin{equation}
    C_p = \frac{|\Delta \mathcal{M}^{sim} < -\tau|}{N}
\end{equation}

\noindent
For the simulation confusion rate, we consider both $\Delta \mathcal{M} > \tau$ and $\Delta \mathcal{M}^{sim} > \tau$. The reason for this is that if $\Delta \mathcal{M}^{sim} > \tau$ then the simulation performs worse under the replay policy, without accounting for the bias of the ground truth policy. However, If a dynamic agent collides with the ego vehicle, it will not be reflected in $\Delta \mathcal{M}^{sim}$, which is why we consider $\Delta \mathcal{M} > \tau$ as well. While $\Delta \mathcal{M}$ is biased due to having ground truth data, it is biased towards $\Delta \mathcal{M} < \tau$, so the union between both metric sets removes this bias. For the policy confusion rate, we consider $\Delta \mathcal{M}^{sim} < -\tau$ which indicates that the simulation performs better after forcing the ego to replay its trajectory, suggesting that the original policy is confusing the traffic simulation.

\subsection{Causal Agent Aware Metametric}
\noindent
Since we are interested in evaluating traffic simulations from the perspective of policy training, we want to ensure that the metametric evaluation contains all agents that are causal to the ego vehicle. In the WOSAC challenge, only a subset of the agents in the traffic simulation need evaluation, and this subset does not always include all the causal agents to the ego vehicle. Using the dataset provided by Roelofs et al. \cite{CausalAgents} we extend the set of evaluated objects to include all the causal agents. We find that this is an important addition as when we compute the delta metrics on the set of causal agents, $\Delta \mathcal{M}_{causal}$ and $\Delta \mathcal{M}_{causal}^{sim}$, scenarios that would have otherwise performed well are revealed to have poorer simulation performance. Thus, all further references to $\Delta \mathcal{M}$ and $\Delta \mathcal{M}^{sim}$ will be computed over $\mathcal{D} = \mathcal{D}_{eval} \cup \mathcal{D}_{causal}$ unless otherwise specified. 

\begin{figure*}[!ht]
    \centering
    \includegraphics[scale=0.25]{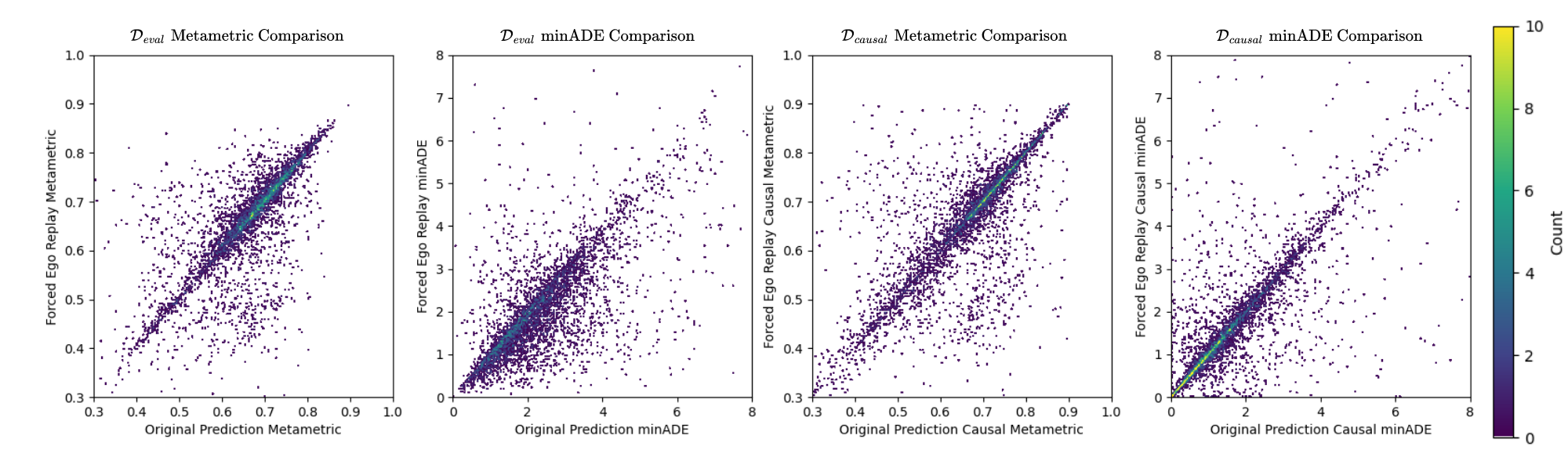}
    \caption{We plot the per-scenario forced replay versus original prediction metametric and minADE for the TrafficBots V1.5 world model. A majority of scenarios show a minimal deviation from $y=x$ indicating minimal impact of an uncontrollable ego agent for these examples. However, there is a large tail of scenarios that show a significant change in both the metametric ($\Delta \mathcal{M}^{sim} > 0.05$) and minADE metrics.}
    \label{fig:scatter}

    \vspace{-4mm}
\end{figure*}

\subsection{Control Dropout for World Model Robustness}
\noindent
The standard approach to using a world model to predict traffic futures is to predict all objects simultaneously using the same model. As such, there is a potential for a world model to become biased towards creating favorable futures for all agents, which can cause the failure case depicted in Figure \ref{fig:sim_agent_compare} when the model does not have complete control of the entire simulation. To resolve this issue, during training, we randomly sample agents with some probability $p_{drop}$ for which we use the agent GT policy instead of $q(o^{sim}_t|o^c_{< t})$ to update the future states of the agent. We call this training strategy control dropout, and it allows the world model to become more robust when agent observations are out of distribution on the world model prediction.

\vspace{-5mm}
\begin{figure}[!h]
    \centering
    \includegraphics[width=0.95\columnwidth]{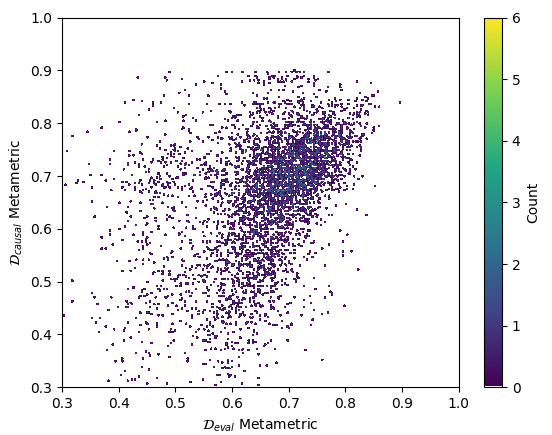}
    \caption{TrafficBots V1.5 per-scenario metametric on $\mathcal{D}_{eval}$ versus $\mathcal{D}_{causal}$}
    \label{fig:domain_scatter}
    \vspace{-5mm}
\end{figure}

\section{Experiments \label{s4}}
\subsection{Model Sensitivity to Uncontrollable Objects}
\noindent
Inspired by Roelofs et al. \cite{CausalAgents}, to understand model sensitivity to uncontrollable objects on a per-scenario level, Figure \ref{fig:scatter} plots the simulation minADE and metametrics \cite{WOMD} for $q^{sim}$ versus $\hat{q}^{sim}$ across the 32 predicted futures using the TrafficBots V1.5 world model. Recall that predictions made in $q^{sim}$ and $\hat{q}^{sim}$  have observations that include the ego vehicle, however, the ego vehicle is not included in the metametric computation to avoid the bias introduced by $\hat{q}^{sim}$ when the ego vehicle follows the replay policy. The figure includes plots for the different evaluation domains $\mathcal{D}_{eval}$, and $\mathcal{D}_{causal}$. Due to the high computational requirements for evaluating $\mathcal{M}$, we perform all evaluations on 10\% (4393 samples) from the WOMD evaluation dataset. While a majority of scenarios show minimal change to the presence of uncontrollable objects in the scene, as indicated by the large cluster on the line $y=x$, there is a large number of outliers indicating that an uncontrollable object creates confusion in the world model prediction. \\

\vspace{-5mm}
\subsection{Evaluation Domain Analysis}
\noindent
Comparing the evaluations on $D_{eval}$ and $D_{causal}$ reveals some interesting discrepancies between the domains. Figure \ref{fig:domain_scatter} shows a scatter plot of the per-scenario metametric on the standard evaluation domain versus the causal vehicle evaluation domain. There is some clustering along $y = x$, however, there is a large variance. This indicates that just because a simulator performs well on the standard evaluation domain does not mean that the agents important to the ego vehicle are being simulated well and vice-versa. \\

\noindent
Inspecting the simulation delta metric, $\Delta \mathcal{M}^{sim}$, reveals that causal agent simulations are more sensitive to uncontrollable objects than the standard evaluation domain. Specifically, $\Delta \mathcal{M}^{sim}_{eval} = 0.025$ while $\Delta \mathcal{M}^{sim}_{causal} = 0.042$, suggesting that causal agents are 68\% more sensitive to an uncontrollable ego agent as compared to agents in $D_{eval}$ for the TrafficBots V1.5 model. This makes sense as causality relationships are usually bidirectional, meaning that there is a high likelihood that the ego agent is also causal to the agents in $D_{causal}$. Thus, if the overall simulation is sensitive to uncontrollable objects, then the agents most likely to be impacted are those that have a causal relationship with the uncontrollable objects. Despite this, there is more dense clustering around $y=x$ for minADE metrics on $D_{causal}$ than $D_{eval}$. This suggests that the deviation in the delta metametric that is observed on $D_{causal}$ is more likely caused by deviations in the WOSAC interaction metrics that affect ADE metrics less, rather than the map-based or kinematic metrics, as seen in the motivating case illustrated in figure \ref{fig:sim_agent_compare}. 

\vspace{-1mm}
\subsection{Policy Aware Metametric Analysis}
\noindent
Considering the standard, causal, and combined evaluation domains, in Table \ref{tab:metric_results} we inspect the failure cases of the TrafficBots V1.5 world model based on $\Delta \mathcal{M}$ and $\Delta \mathcal{M}^{sim}$. 

\begin{table}[!ht]
    \centering
    \caption{World Model Failure Case Analysis}
    \resizebox{\columnwidth}{!}{%
    \begin{tabular}{c|c|c|c|c|c}
        \hline\hline
          \Gape[0.25cm][0.25cm]{Domain} & Threshold & $\Delta \mathcal{M}_i > \tau$ & $\Delta \mathcal{M}^{sim}_i > \tau$  & $\Delta \mathcal{M}_i \cup \Delta \mathcal{M}^{sim}_i > \tau$ & $\Delta \mathcal{M}^{sim}_i < -\tau$ \\
          \hline
          \multirow{2}{4em}{$\mathcal{D}_{eval}$} & $\tau = 0.05$ & 0.119 & 0.076 & 0.167 & 0.064 \\
          & $\tau = 0.035$ & 0.138 & 0.090 & 0.193 & 0.077 \\
          \hline
          \multirow{2}{4em}{$\mathcal{D}_{causal}$} & $\tau = 0.05$ & 0.125 & 0.115 & 0.184 & 0.085 \\
          & $\tau = 0.035$ & 0.161 & 0.131 & 0.224 & 0.104 \\
          \hline
          \multirow{2}{4em}{$\mathcal{D}_{eval} \cup \mathcal{D}_{causal}$} & $\tau = 0.05$ & 0.181 & 0.142 & 0.248 & 0.109 \\
          & $\tau = 0.035$ & 0.222 & 0.161 & 0.295 & 0.132 \\
        \hline\hline
    \end{tabular}%
    }
    \vspace{1mm}
    \begin{flushleft}
    Table \ref{tab:metric_results}. TrafficBots V1.5 failure analysis using new delta metrics on standard and causal evaluation domains.
    \end{flushleft}
    \label{tab:metric_results}
    \vspace{-3mm}
\end{table}

\begin{figure}[!ht]
    \centering
    \includegraphics[width=\columnwidth]{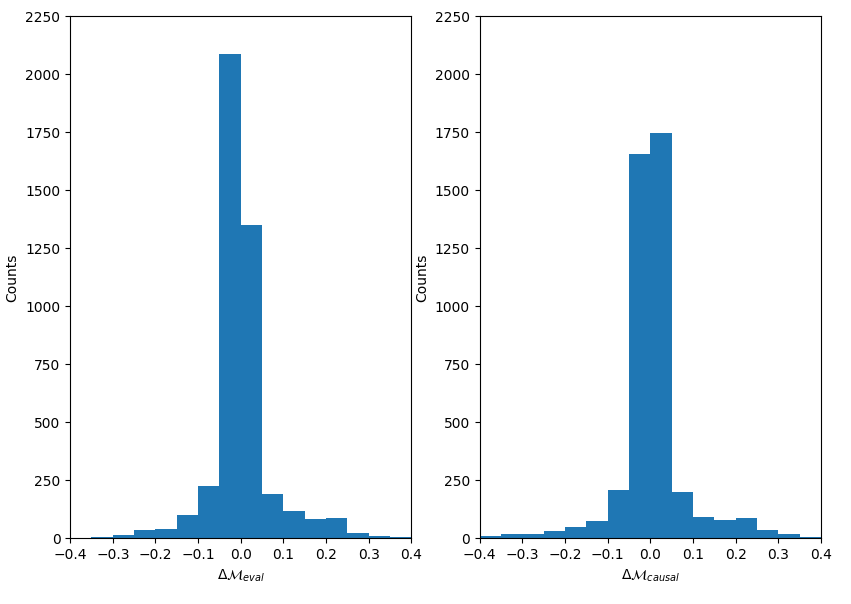}
    \caption{Distribution of $\Delta \mathcal{M}$ over 4393 evaluation scenarios. Most cases have $\Delta \mathcal{M} \approx 0$, but the large tail of the distribution for scenarios with $|\Delta \mathcal{M}| > 0.05$ indicates potential for high sensitivity in the world model traffic simulation.}
    \label{fig:meta_histogram}
\end{figure}

\noindent
We consider two thresholds; $\tau = 0.035$, which is one standard deviation in the metametric on the WOSAC leaderboard at the time of writing, and $\tau = 0.05$, which is the difference between the leading model and TrafficBots V1.5 model on the WOSAC leaderboard at the time of writing. Recall the definition the delta metric from Eq. \ref{eq:delta_metric}. The first two columns of the table evaluate the frequency of errors caused by the traffic simulation. If $\Delta \mathcal{M} > \tau$ then the simulation is confused by the use of the GT policy. If $\Delta \mathcal{M}^{sim} > \tau$ then the simulation is confused by the GT policy, even when removing the GT policy from the evaluation. The third table column aggregates the previous two failure cases to evaluate the total simulation confusion rate, $\mathcal{C}_s$. The final column evaluates the frequency of errors caused by the ego agent policy, this the policy confusion rate, $\mathcal{C}_p$. While we choose to consider thresholds of $\tau = 0.035$ and $\tau = 0.05$, there is a long tail of scenarios which have much larger deviations in the new delta metrics. Figure \ref{fig:meta_histogram} depicts a histogram of $\Delta \mathcal{M}$ over all evaluated scenarios from both the standard evaluation domain and the causal agents evaluation domain. \\

\noindent
Following the same procedure for evaluation, Table \ref{tab:model_comparison} analyzes the metametrics and confusion rates for the TrafficBots V1.5, GUMP, and SMART world models on the WOMD evaluation dataset over $\mathcal{D}_{eval}$. The threshold value used to compute the confusion rates is $\tau = 0.05$ and to ensure fairness between the evaluation of all world models, each model was trained for 10 epochs on only the WOMD data. Interestingly, although GUMP performs better on the standard WOSAC realism metric, it seems more sensitive to uncontrollable dynamic objects. Furthermore, it can be seen that the presence of uncontrollable objects does not impact the SMART world model. This is because SMART is trained in an open-loop manner, and thus is already robust to the presence of uncontrollable objects. This is expected as it is only models trained autoregressively with full control over all agents that develop a sensitivity to uncontrollable objects, which is the case for most state-of-the-art world models. Despite this, the simulation and policy confusion metrics are still useful in indicating whether poor performance in a scenario is due to the world model rollout or the ego policy. 

\vspace{-3mm}
\begin{table}[!h]
    \centering
    \caption{World Model Metametric Analysis}
    \resizebox{\columnwidth}{!}{%
    \begin{tabular}{c|c|c|c|c|c}
        \hline\hline
          \Gape[0.25cm][0.25cm]{Model} & $\mathcal{M}_{eval}$ & $\Delta \mathcal{M}_{eval}$ & $\Delta \mathcal{M}_{eval}^{sim}$ & $\mathcal{C}_s$ & $\mathcal{C}_p$ \\
          \hline
          TrafficBots V1.5 \cite{Trafficbots} & 0.662 & 0.039 & 0.025 & 0.167 & 0.064 \\
          \hline
          GUMP \cite{GUMP} & 0.702 & 0.079 & 0.038  & 0.356 & 0.063 \\
          \hline
          SMART \cite{SMART} & 0.746 & -0.003 & 0.001  & 0.006 & 0.002 \\
        \hline\hline
    \end{tabular}%
    }
    \vspace{1mm}
    \begin{flushleft}
    Table \ref{tab:model_comparison}. Comparison of the WOSAC metametric with the newly proposed delta metrics and confusion rates. By itself, the original metametric does not describe the performance of the model well from the perspective of the ego vehicle, however, the new metrics can be used to identify whether the ego policy or the traffic simulation is more at fault for a poorly performing episode. 
    \end{flushleft}
    \label{tab:model_comparison}
    \vspace{-3mm}
\end{table}

\subsection{Control Dropout for Simulation Robustness}
\noindent
To improve robustness to uncontrollable objects, the control dropout training strategy was employed during autoregressive training. Table \ref{tab:model_comparison_dropout} analyzes the metametrics and confusion rates for the TrafficBots V1.5 world model after training for 10 epochs using a control dropout with $p_{drop} = 0.1$. As seen from the results, while the standard metametric is relatively unaffected, the control dropout training strategy decreases $\Delta \mathcal{M}$ indicating that the model with control dropout is more robust when uncontrollable objects are present in the scenario. This is further bolstered by the decrease in simulation confusion rate, $C_s$, between the default and control dropout models.

\begin{table}[!h]
    \centering
    \caption{Control Dropout Metametric Analysis}
    \resizebox{\columnwidth}{!}{%
    \begin{tabular}{c|c|c}
        \hline\hline
          \Gape[0.25cm][0.25cm]{Model} & TrafficBots V1.5 & TrafficBots V1.5 + drop \\
          \hline
          \Gape[0.15cm][0.15cm]{$\mathcal{M}_{eval}$} & 0.662 & 0.655 \\
          \hline
          \Gape[0.15cm][0.15cm]{$\Delta \mathcal{M}_{eval}$} & 0.039 & 0.023 \\
          \hline
          \Gape[0.15cm][0.15cm]{$\Delta \mathcal{M}_{eval}^{sim}$} & 0.025 & 0.007  \\
          \hline
          \Gape[0.1cm][0.1cm]{$\mathcal{C}_s$} & 0.167 & 0.072 \\
          \hline
          \Gape[0.1cm][0.1cm]{$\mathcal{C}_p$} & 0.064 & 0.003 \\
        \hline\hline
    \end{tabular}%
    }
    \vspace{1mm}
    \begin{flushleft}
    Table \ref{tab:model_comparison_dropout}. Model comparison of TrafficBots V1.5 trained with and without control dropout. While the metametric is not improved by training with control dropout, the delta metametrics are significantly reduced, leading to a drop in simulation confusion. 
    \end{flushleft}
    \label{tab:model_comparison_dropout}
\end{table}

\vspace{-3mm}
\section{Discussion \& Interpretation \label{s5}}
\subsection{World Model Sensitivity}
\noindent
Our results show that world models trained autoregressively to predict futures for all objects are sensitive to the presence of uncontrollable objects in the environment. Models that are trained in an autoregressive closed-loop setting like GUMP and TrafficBots V1.5 are more sensitive than models trained with open-loop objectives like SMART. By introducing control dropout in autoregressive training, we partially shift the distribution of observations towards that of open-loop training and force the autoregressive model to become more robust when it must make predictions in environments with uncontrollable objects.  \\

\subsection{Importance of Causal Agent Evaluation}
\noindent
Figure \ref{fig:domain_scatter} reveals a big discrepancy in the performance of traffic rollouts depending on which domain of agents the simulation is evaluated on. We believe that if we want to train an autonomous driving planning agent inside the world model simulation, the simulation of agents that are causal to the ego vehicle throughout the simulation must be realistic. It can be expected that if a planning agent is penalized for mistakes made by the traffic simulation, then the behavior of the planning agent will become less well-defined as it may attempt to avoid interactive scenarios or drive overly cautious.    

\section{Conclusions \label{s6}}
\noindent
In this work we propose new metrics for evaluating world models as data-driven traffic simulators that provides better insight on the performance of separate ego policy and traffic simulator performance compared to the default WOSAC metametric. Different evaluation domains are also explored, and it is revealed that traffic simulation rollouts of causal agents often diverge in quality compared to simulation rollouts conducted on the standard WOSAC evaluation domain.  Furthermore, we demonstrate that existing state-of-the-art world models are sensitive to perturbations that introduce uncontrollable dynamic objects in the scene if these models were trained autoregressively. To address this problem of model sensitivity, we introduce a novel training strategy called control dropout which successfully makes world model traffic simulations more robust when only part of the scenario can be simulated.

\section*{ACKNOWLEDGEMENT}
\noindent
The authors would like to thank Ehsan Ahmadi and Yibo Liu for the constructive discussions and insight.

\bibliographystyle{IEEEtran.bst}

\typeout{}
\bibliography{reference.bib}

\begin{thebibliography}{10}
\providecommand{\url}[1]{#1}
\csname url@rmstyle\endcsname
\providecommand{\newblock}{\relax}
\providecommand{\bibinfo}[2]{#2}
\providecommand\BIBentrySTDinterwordspacing{\spaceskip=0pt\relax}
\providecommand\BIBentryALTinterwordstretchfactor{4}
\providecommand\BIBentryALTinterwordspacing{\spaceskip=\fontdimen2\font plus
\BIBentryALTinterwordstretchfactor\fontdimen3\font minus \fontdimen4\font\relax}
\providecommand\BIBforeignlanguage[2]{{%
\expandafter\ifx\csname l@#1\endcsname\relax
\typeout{** WARNING: IEEEtran.bst: No hyphenation pattern has been}%
\typeout{** loaded for the language `#1'. Using the pattern for}%
\typeout{** the default language instead.}%
\else
\language=\csname l@#1\endcsname
\fi
#2}}

\bibitem{GameFormer}
\BIBentryALTinterwordspacing
Z.~Huang, H.~Liu, and C.~Lv, ``Gameformer: Game-theoretic modeling and learning of transformer-based interactive prediction and planning for autonomous driving,'' 2023. [Online]. Available: \url{https://arxiv.org/abs/2303.05760}
\BIBentrySTDinterwordspacing

\bibitem{PDM}
\BIBentryALTinterwordspacing
D.~Dauner, M.~Hallgarten, A.~Geiger, and K.~Chitta, ``Parting with misconceptions about learning-based vehicle motion planning,'' in \emph{7th Annual Conference on Robot Learning}, 2023. [Online]. Available: \url{https://openreview.net/forum?id=o82EXEK5hu6}
\BIBentrySTDinterwordspacing

\bibitem{NuPlan2nd}
\BIBentryALTinterwordspacing
Y.~Hu, K.~Li, P.~Liang, J.~Qian, Z.~Yang, H.~Zhang, W.~Shao, Z.~Ding, W.~Xu, and Q.~Liu, ``Imitation with spatial-temporal heatmap: 2nd place solution for nuplan challenge,'' 2023. [Online]. Available: \url{https://arxiv.org/abs/2306.15700}
\BIBentrySTDinterwordspacing

\bibitem{Trafficbots}
Z.~Zhang, A.~Liniger, D.~Dai, F.~Yu, and L.~Van~Gool, ``Trafficbots: Towards world models for autonomous driving simulation and motion prediction,'' in \emph{2023 IEEE International Conference on Robotics and Automation (ICRA)}, 2023, pp. 1522--1529.

\bibitem{SMART}
\BIBentryALTinterwordspacing
W.~Wu, X.~Feng, Z.~Gao, and Y.~Kan, ``Smart: Scalable multi-agent real-time simulation via next-token prediction,'' 2024. [Online]. Available: \url{https://arxiv.org/abs/2405.15677}
\BIBentrySTDinterwordspacing

\bibitem{WOMD}
S.~Ettinger, S.~Cheng, B.~Caine, C.~Liu, H.~Zhao, S.~Pradhan, Y.~Chai, B.~Sapp, C.~Qi, Y.~Zhou, Z.~Yang, A.~Chouard, P.~Sun, J.~Ngiam, V.~Vasudevan, A.~McCauley, J.~Shlens, and D.~Anguelov, ``Large scale interactive motion forecasting for autonomous driving : The waymo open motion dataset,'' in \emph{2021 IEEE/CVF International Conference on Computer Vision (ICCV)}, 2021, pp. 9690--9699.

\bibitem{nuplan}
K.~T. e.~a. H.~Caesar, J.~Kabzan, ``Nuplan: A closed-loop ml-based planning benchmark for autonomous vehicles,'' in \emph{CVPR ADP3 workshop}, 2021.

\bibitem{CARLA}
A.~Dosovitskiy, G.~Ros, F.~Codevilla, A.~Lopez, and V.~Koltun, ``{CARLA}: {An} open urban driving simulator,'' in \emph{Proceedings of the 1st Annual Conference on Robot Learning}, 2017, pp. 1--16.

\bibitem{SMARTS}
\BIBentryALTinterwordspacing
M.~Zhou, J.~Luo, J.~Villella, Y.~Yang, D.~Rusu, J.~Miao, W.~Zhang, M.~Alban, I.~Fadakar, Z.~Chen, A.~C. Huang, Y.~Wen, K.~Hassanzadeh, D.~Graves, D.~Chen, Z.~Zhu, N.~Nguyen, M.~Elsayed, K.~Shao, S.~Ahilan, B.~Zhang, J.~Wu, Z.~Fu, K.~Rezaee, P.~Yadmellat, M.~Rohani, N.~P. Nieves, Y.~Ni, S.~Banijamali, A.~C. Rivers, Z.~Tian, D.~Palenicek, H.~bou Ammar, H.~Zhang, W.~Liu, J.~Hao, and J.~Wang, ``Smarts: Scalable multi-agent reinforcement learning training school for autonomous driving,'' 11 2020. [Online]. Available: \url{https://arxiv.org/abs/2010.09776}
\BIBentrySTDinterwordspacing

\bibitem{metadrive}
Q.~Li, Z.~Peng, L.~Feng, Q.~Zhang, Z.~Xue, and B.~Zhou, ``Metadrive: Composing diverse driving scenarios for generalizable reinforcement learning,'' \emph{IEEE Transactions on Pattern Analysis and Machine Intelligence}, 2022.

\bibitem{sim2real}
W.~Zhao, J.~P. Queralta, and T.~Westerlund, ``Sim-to-real transfer in deep reinforcement learning for robotics: a survey,'' in \emph{2020 IEEE Symposium Series on Computational Intelligence (SSCI)}, 2020, pp. 737--744.

\bibitem{GUMP}
\BIBentryALTinterwordspacing
Y.~Hu, S.~Chai, Z.~Yang, J.~Qian, K.~Li, W.~Shao, H.~Zhang, W.~Xu, and Q.~Liu, ``Solving motion planning tasks with a scalable generative model,'' 2024. [Online]. Available: \url{https://arxiv.org/abs/2407.02797}
\BIBentrySTDinterwordspacing

\bibitem{WOSAC}
\BIBentryALTinterwordspacing
N.~Montali, J.~Lambert, P.~Mougin, A.~Kuefler, N.~Rhinehart, M.~Li, C.~Gulino, T.~Emrich, Z.~Yang, S.~Whiteson, B.~White, and D.~Anguelov, ``The waymo open sim agents challenge,'' in \emph{Advances in Neural Information Processing Systems}, A.~Oh, T.~Naumann, A.~Globerson, K.~Saenko, M.~Hardt, and S.~Levine, Eds., vol.~36.\hskip 1em plus 0.5em minus 0.4em\relax Curran Associates, Inc., 2023, pp. 59\,151--59\,171. [Online]. Available: \url{https://proceedings.neurips.cc/paper_files/paper/2023/file/b96ce67b2f2d45e4ab315e13a6b5b9c5-Paper-Datasets_and_Benchmarks.pdf}
\BIBentrySTDinterwordspacing

\bibitem{DataDrivenSim}
T.-H. Wang, A.~Amini, W.~Schwarting, I.~Gilitschenski, S.~Karaman, and D.~Rus, ``Learning interactive driving policies via data-driven simulation,'' in \emph{2022 International Conference on Robotics and Automation (ICRA)}, 2022, pp. 7745--7752.

\bibitem{KIGRAS}
\BIBentryALTinterwordspacing
J.~Zhao, J.~Zhuang, Q.~Zhou, T.~Ban, Z.~Xu, H.~Zhou, J.~Wang, G.~Wang, Z.~Li, and B.~Li, ``Kigras: Kinematic-driven generative model for realistic agent simulation,'' 2024. [Online]. Available: \url{https://arxiv.org/abs/2407.12940}
\BIBentrySTDinterwordspacing

\bibitem{VRD}
H.~Schofield, H.~Mirkhani, M.~Elmahgiubi, K.~Rezaee, and J.~Shan, ``Vectorized representation dreamer (vrd): Dreaming-assisted multi-agent motion forecasting,'' in \emph{2024 IEEE Intelligent Vehicles Symposium (IV)}, 2024, pp. 2012--2017.

\bibitem{MILE}
\BIBentryALTinterwordspacing
A.~Hu, G.~Corrado, N.~Griffiths, Z.~Murez, C.~Gurau, H.~Yeo, A.~Kendall, R.~Cipolla, and J.~Shotton, ``Model-based imitation learning for urban driving,'' in \emph{Advances in Neural Information Processing Systems}, S.~Koyejo, S.~Mohamed, A.~Agarwal, D.~Belgrave, K.~Cho, and A.~Oh, Eds., vol.~35.\hskip 1em plus 0.5em minus 0.4em\relax Curran Associates, Inc., 2022, pp. 20\,703--20\,716. [Online]. Available: \url{https://proceedings.neurips.cc/paper_files/paper/2022/file/827cb489449ea216e4a257c47e407d18-Paper-Conference.pdf}
\BIBentrySTDinterwordspacing

\bibitem{Dream2Control}
\BIBentryALTinterwordspacing
D.~Hafner, T.~Lillicrap, J.~Ba, and M.~Norouzi, ``Dream to control: Learning behaviors by latent imagination,'' in \emph{International Conference on Learning Representations}, 2020. [Online]. Available: \url{https://openreview.net/forum?id=S1lOTC4tDS}
\BIBentrySTDinterwordspacing

\bibitem{DreamerV2}
\BIBentryALTinterwordspacing
D.~Hafner, T.~P. Lillicrap, M.~Norouzi, and J.~Ba, ``Mastering atari with discrete world models,'' in \emph{International Conference on Learning Representations}, 2021. [Online]. Available: \url{https://openreview.net/forum?id=0oabwyZbOu}
\BIBentrySTDinterwordspacing

\bibitem{Zhao_TNT}
H.~Zhao, J.~Gao, T.~Lan, C.~Sun, B.~Sapp, B.~Varadarajan, Y.~Shen, Y.~Shen, Y.~Chai, C.~Schmid, C.~Li, and D.~Anguelov, ``{TNT: Target-driven Trajectory Prediction},'' in \emph{Proceedings of the 2020 Conference on Robot Learning}, ser. Proceedings of Machine Learning Research, vol. 155.\hskip 1em plus 0.5em minus 0.4em\relax PMLR, 16--18 Nov 2021, pp. 895--904.

\bibitem{SimRL}
B.~Osinski, A.~Jakubowski, P.~Ziecina, P.~Milos, C.~Galias, S.~Homoceanu, and H.~Michalewski, ``Simulation-based reinforcement learning for real-world autonomous driving,'' in \emph{2020 IEEE International Conference on Robotics and Automation (ICRA)}, 2020, pp. 6411--6418.

\bibitem{DQNPlanning}
M.~Ahmed, C.~P. Lim, and S.~Nahavandi, ``A deep q-network reinforcement learning-based model for autonomous driving,'' in \emph{2021 IEEE International Conference on Systems, Man, and Cybernetics (SMC)}, 2021, pp. 739--744.

\bibitem{ImitationNotEnough}
Y.~Lu, J.~Fu, G.~Tucker, X.~Pan, E.~Bronstein, R.~Roelofs, B.~Sapp, B.~White, A.~Faust, S.~Whiteson, D.~Anguelov, and S.~Levine, ``{Imitation Is Not Enough: Robustifying Imitation with Reinforcement Learning for Challenging Driving Scenarios},'' in \emph{2023 IEEE/RSJ International Conference on Intelligent Robots and Systems (IROS)}, 2023, pp. 7553--7560.

\bibitem{HREIL}
Z.~Cao, E.~Biyik, W.~Wang, A.~Raventos, A.~Gaidon, G.~Rosman, and D.~Sadigh, ``{Reinforcement Learning based Control of Imitative Policies for Near-Accident Driving},'' in \emph{Proceedings of Robotics: Science and Systems}, Corvalis, Oregon, USA, July 2020.

\bibitem{DAGGER}
\BIBentryALTinterwordspacing
S.~Ross, G.~J. Gordon, and J.~A. Bagnell, ``A reduction of imitation learning and structured prediction to no-regret online learning,'' 2011. [Online]. Available: \url{https://arxiv.org/abs/1011.0686}
\BIBentrySTDinterwordspacing

\bibitem{CausalRepresentation}
B.~Schölkopf, F.~Locatello, S.~Bauer, N.~R. Ke, N.~Kalchbrenner, A.~Goyal, and Y.~Bengio, ``Toward causal representation learning,'' \emph{Proceedings of the IEEE}, vol. 109, no.~5, pp. 612--634, 2021.

\bibitem{CausalAnnotations}
V.~Ramanishka, Y.-T. Chen, T.~Misu, and K.~Saenko, ``Toward driving scene understanding: A dataset for learning driver behavior and causal reasoning,'' in \emph{2018 IEEE/CVF Conference on Computer Vision and Pattern Recognition}, 2018, pp. 7699--7707.

\bibitem{CausalRepresentationDomainAdaption}
S.~Yang, K.~Yu, F.~Cao, L.~Liu, H.~Wang, and J.~Li, ``Learning causal representations for robust domain adaptation,'' \emph{IEEE Transactions on Knowledge and Data Engineering}, vol.~35, no.~3, pp. 2750--2764, 2023.

\bibitem{CausalOOD}
\BIBentryALTinterwordspacing
C.~Liu, X.~Sun, J.~Wang, H.~Tang, T.~Li, T.~Qin, W.~Chen, and T.-Y. Liu, ``Learning causal semantic representation for out-of-distribution prediction,'' in \emph{Advances in Neural Information Processing Systems}, M.~Ranzato, A.~Beygelzimer, Y.~Dauphin, P.~Liang, and J.~W. Vaughan, Eds., vol.~34.\hskip 1em plus 0.5em minus 0.4em\relax Curran Associates, Inc., 2021, pp. 6155--6170. [Online]. Available: \url{https://proceedings.neurips.cc/paper_files/paper/2021/file/310614fca8fb8e5491295336298c340f-Paper.pdf}
\BIBentrySTDinterwordspacing

\bibitem{OODCausal}
\BIBentryALTinterwordspacing
S.~S.~G. Bagi, Z.~Gharaee, O.~Schulte, and M.~Crowley, ``Generative causal representation learning for out-of-distribution motion forecasting,'' 2023. [Online]. Available: \url{https://arxiv.org/abs/2302.08635}
\BIBentrySTDinterwordspacing

\bibitem{Sim2RealCausal}
\BIBentryALTinterwordspacing
Y.~Liu, A.~Rahimi, P.-C. Luan, F.~Rajič, and A.~Alahi, ``Sim-to-real causal transfer: A metric learning approach to causally-aware interaction representations,'' 2023. [Online]. Available: \url{https://arxiv.org/abs/2312.04540}
\BIBentrySTDinterwordspacing

\bibitem{CausalAgents}
L.~Sun, R.~Roelofs, B.~Caine, K.~S. Refaat, B.~Sapp, S.~Ettinger, and W.~Chai, ``Causalagents: A robustness benchmark for motion forecasting,'' in \emph{2024 IEEE International Conference on Robotics and Automation (ICRA)}, 2024, pp. 6820--6827.

\end{thebibliography}

\end{document}